%% file: main.tex
\documentclass{article}
\usepackage{paper_template,times}
\makeatletter
\ClearShipoutPicture
\def\@maketitle{\vbox{\hsize\textwidth
{\LARGE\sc \@title\par}
\lhead{\textit{Preprint}}
\def\And{\end{tabular}\hfil\linebreak[0]\hfil
\begin{tabular}[t]{l}\bf\rule{\z@}{24pt}\ignorespaces}%
\def\AND{\end{tabular}\hfil\linebreak[4]\hfil
\begin{tabular}[t]{l}\bf\rule{\z@}{24pt}\ignorespaces}%
\begin{tabular}[t]{l}\bf\rule{\z@}{24pt}\@author\end{tabular}%
\vskip 0.3in minus 0.1in}}
\makeatother
\input{math_commands.tex}

\usepackage{graphicx}
\usepackage{hyperref}
\usepackage{url}
\usepackage{booktabs}
\usepackage{array}
\usepackage{float}

\newcommand{\localtablecaption}[1]{%
  \refstepcounter{table}%
  {\small\raggedright\noindent\textbf{Table~\thetable:} #1\par}%
  \vspace{2pt}%
}
\newcommand{\localfigurecaption}[1]{%
  \refstepcounter{figure}%
  {\small\raggedright\noindent\textbf{Figure~\thefigure:} #1\par}%
  \vspace{2pt}%
}
\title{Planning in the Backbone: DiffAdapterVLA for Native Continuous Trajectory Generation with Driving VLMs}

\author{%
\textbf{Changxin Lu$^{1,2}$\thanks{Work done during an internship at Dongfeng Research \& Development Institute.} \qquad Xiaoliang Meng$^{1}$\thanks{Corresponding author.}}\\
\textbf{Yu Wu$^{2}$\thanks{Project leader.} \quad Rui Huang$^{2}$ \quad Honglin Li$^{2}$ \quad Tao Chen$^{2}$ \quad Kaixuan Zhou$^{2}$ \quad Yadong Shao$^{2}$}\\
$^{1}$School of Remote Sensing and Information Engineering, Wuhan University\\
$^{2}$Dongfeng Research \& Development Institute\\
\texttt{\{2021302131130,xmeng\}@whu.edu.cn}\\
\texttt{\{dfrd-wuyu,tc-huangrui,lihongl,chentao,shaoyadong\}@dfmc.com.cn}\\
\texttt{triumphzhou@163.com}%
}

\begin{document}
\raggedbottom
\maketitle

\begin{abstract}
Pretrained driving vision-language models (VLMs) integrate visual, route,
language, and driving context into rich driving priors, yet their representation
objectives remain separated from continuous driving planning. Existing methods
typically begin trajectory generation only after the VLM has formed a final
condition, leaving depth-wise condition computation outside the stepwise
formation of trajectory state. We introduce \textbf{DiffAdapterVLA}, which
realizes \emph{Planning in the Backbone}: it injects explicit trajectory tokens
into selected VLM late layers, bringing trajectory state into backbone forward
computation, where it co-evolves with driving conditions at different depths.
Lightweight layer-wise DiffAdapters organize this computation into recursive
trajectory refinement, while asymmetric joint attention preserves directed
guidance from the condition stream to trajectory planning. By placing planning
within existing backbone computation rather than relying on an independent
trajectory planner, DiffAdapterVLA adapts only lightweight trajectory modules
to turn existing driving priors into efficient continuous planning capability.
NAVSIM results show that it achieves high-quality closed-loop planning with low
end-to-end latency using few trainable parameters, and demonstrate that jointly
evolving trajectory state and depth-wise driving conditions in VLM late-layer
computation effectively realizes continuous trajectory planning.
\end{abstract}

\section{Introduction}

As autonomous driving moves toward open and dynamic traffic, end-to-end (E2E)
planning has become central to generating safe, executable, and
interaction-aware future motion \citep{chen2024e2esurvey}. By unifying
perception, scene modeling, and motion planning, E2E systems have advanced
through multimodal inputs, temporal modeling, and closed-loop evaluation
\citep{prakash2021transfuser,hu2023uniad,jiang2023vad,caesar2021nuplan,dauner2024navsim}.
Driving vision-language models (VLMs) further integrate visual, linguistic, and
driving knowledge, extending conventional E2E systems toward
vision-language-action (VLA) models. As these models produce increasingly rich
driving representations, a central challenge is to convert them into precise
and efficient continuous motion
\citep{gao2024foundationad,zhou2024vlmsurvey,sima2024drivelm,shao2024lmdrive,tian2024drivevlm,zhou2026opendrivevla}.

Yet rich pretrained representations do not naturally translate into reliable
continuous planning. This gap stems from different modeling objectives: general
VLMs learn visual-language semantics through discrete sequence prediction,
whereas driving requires geometrically precise, temporally continuous, and
dynamically feasible trajectories under real-time constraints. Empirical studies
show that VLM initialization can benefit downstream policies, yet general
vision-language competence remains a poor predictor of control performance
\citep{zhang2026vlm4vla}; driving research likewise repeatedly identifies the
mismatch between semantic reasoning and numerical trajectory spaces
\citep{fu2025orion,li2026recogdrive,wang2026linkvla}. Broader VLA research
addresses the same bottleneck by redesigning continuous-action modeling, action
representations, and action tokenization
\citep{hou2025dita,wen2025diffusionvla,pertsch2025fast}. These efforts establish
the action interface as a central problem in converting pretrained driving
representations into continuous trajectories.

\begin{figure*}[t]
    \centering
    \includegraphics[width=0.90\textwidth]{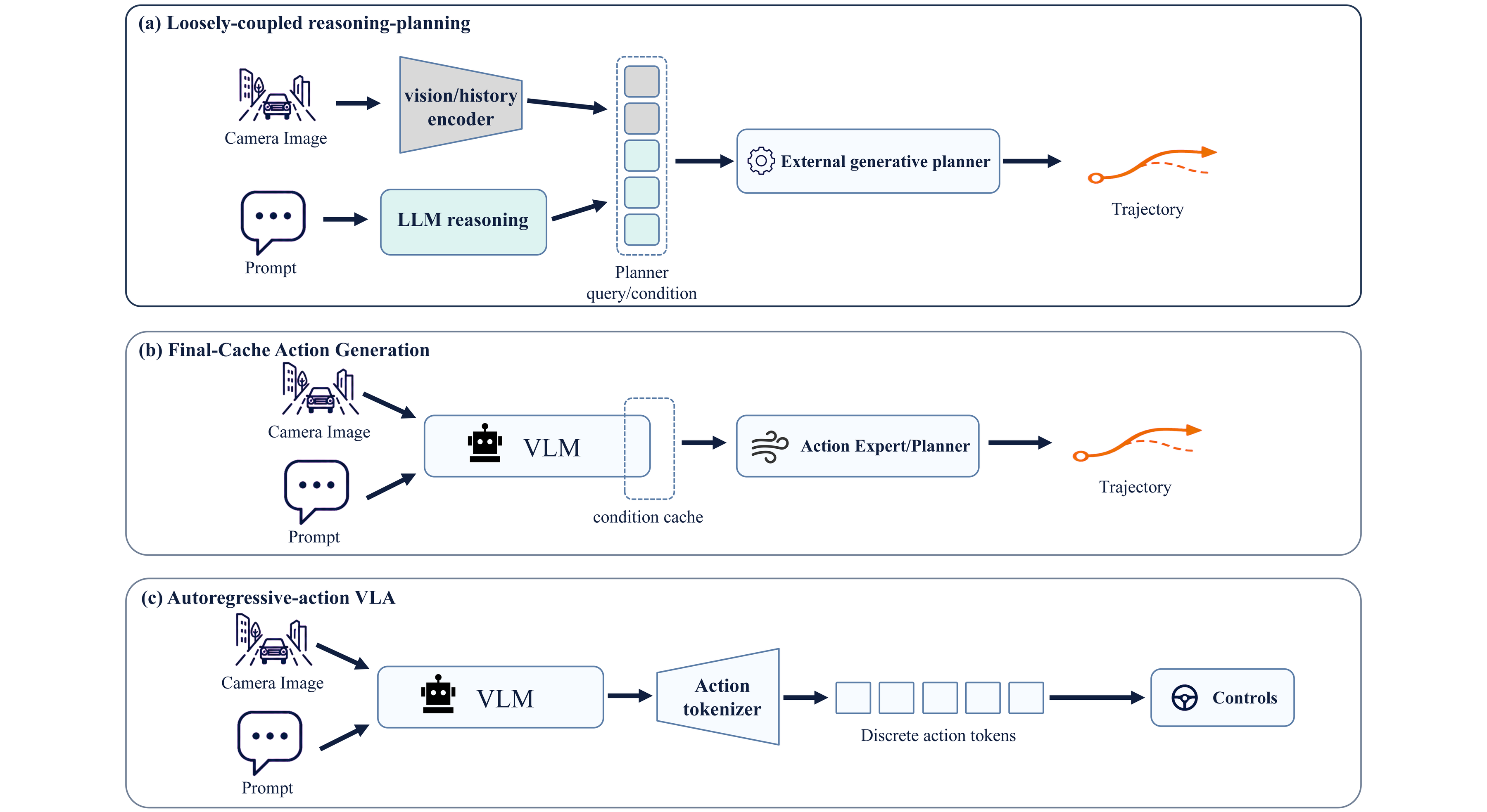}
    \caption{\textbf{Three prevalent interfaces between driving understanding and action generation.}
    (a) Loosely coupled reasoning--planning systems connect separate visual,
    language, and planning modules through intermediate features or cues.
    (b) Final-cache action generation
    passes the condition cache produced by a VLM to an action expert or planner.
    (c) Autoregressive-action VLAs translate VLM representations into discrete
    action tokens before producing controls.}
    \label{fig:action-interfaces}
\end{figure*}

Figure~\ref{fig:action-interfaces} summarizes three representative interfaces
between driving understanding and action generation. Loosely coupled
reasoning--planning systems pass semantic cues from separately encoded visual
and language inputs to an independent trajectory planner, coupling the modules
only through intermediate interfaces
\citep{pan2024vlp,fu2025orion}. Final-condition methods instead pass final-layer
VLM representations, such as hidden states, planning tokens, or condition
caches, to a downstream action expert, trajectory regressor, or diffusion planner
\citep{nvidia2025alpamayo,li2026recogdrive}. Autoregressive VLAs serialize
actions or trajectories as discrete tokens for next-token prediction
\citep{zhou2025autovla,mao2023gptdriver,huang2025drivegpt,rowe2025poutine,
hwang2025emma}. In the first two interfaces, trajectory evolution begins after
VLM condition computation; the third reformulates continuous planning as
discrete generation. In either case, internal condition computation cannot
directly participate in the evolution of a continuous trajectory state.

These interfaces treat trajectory planning as the outcome of scene
understanding, even though driving VLM condition representations are formed
progressively as visual evidence, route information, language, motion history,
and traffic interactions are integrated across layers
\citep{strong2026learning,bevinmllm2024,sima2024drivelm,shao2024lmdrive,
xu2025drivegpt4v2,wang2025omnidrive}. When planning consumes only the final
output, these depth-evolving driving signals cannot directly participate in
forming the trajectory state. To address this limitation, we propose an
interface that brings trajectory formation into the representational process
itself: early layers retain multimodal condition modeling, while an explicit
trajectory state enters selected late layers and evolves jointly with condition
information within the backbone computation. We call this interface backbone-native
continuous planning, or \emph{Planning in the Backbone}. The trajectory is no
longer only the outcome of scene understanding; it takes shape layer by layer
during the latter stages of that computation.

To realize this interface, we introduce DiffAdapterVLA, a lightweight layer-wise
architecture that recursively refines continuous trajectory states within
selected backbone layers, enabling parameter-efficient, low-latency planning
without an external trajectory planner.

Our contributions are as follows:

{\leftskip=1.2em
\noindent\textbullet\ We establish and validate a backbone-native continuous
planning interface in which an explicit trajectory state enters the late-layer
computation of a pretrained driving VLM and progressively forms a continuous
trajectory from driving conditions at different depths.\par}

{\leftskip=1.2em
\noindent\textbullet\ We introduce DiffAdapterVLA, which recursively updates
the trajectory state through lightweight layer-wise DiffAdapters, achieving
high-performance, low-latency continuous planning with few trainable
parameters.\par}

{\leftskip=1.2em
\noindent\textbullet\ We introduce condition-preserving asymmetric joint
attention, which establishes directed information flow between trajectory
denoising and backbone condition computation while preserving the original
condition stream.\par}

\section{Related Work}

\noindent\textbf{E2E Driving and Continuous Trajectory Planning.}
E2E autonomous driving jointly learns perception, scene understanding,
and planning for future trajectories. Multimodal fusion, planning-oriented joint
modeling, and sparse or vectorized scene representations progressively bring
these functions into shared feature spaces
\citep{prakash2021transfuser,hu2023uniad,jiang2023vad,jia2023thinktwice,sun2025sparsedrive,weng2024paradrive}.
Generative and latent-world approaches extend this paradigm through structural
latent modeling, intention-aware world models, and multi-mode planning
distillation \citep{zheng2024genad,zheng2025world4drive,yu2025distilldrive},
while nuPlan, NAVSIM, and pseudo-simulation assess trajectory feasibility and
interaction quality \citep{caesar2021nuplan,dauner2024navsim,cao2025pseudosimulation}.
DiffusionDrive and GoalFlow further generate continuous trajectories with
truncated diffusion and flow matching, respectively
\citep{liao2025diffusiondrive,xing2025goalflow}. Yet, generation generally
remains a head or module after shared scene representations, rather than an
evolving state interleaved with a pretrained driving VLM's late-layer computation.

\noindent\textbf{Driving VLMs and VLAs for Autonomous Planning.} Driving VLMs unify
visual observations, language, route information, and driving knowledge for
semantic understanding and decision making
\citep{sima2024drivelm,shao2024lmdrive,tian2024drivevlm,zhou2026opendrivevla}.
Driving VLAs then connect this representation to planning through reasoning
conditions followed by a planner \citep{pan2024vlp,fu2025orion}, VLM hidden
states, planning tokens, or caches passed to downstream action components
\citep{nvidia2025alpamayo,li2026recogdrive}, or autoregressive discrete action
and trajectory tokens \citep{zhou2025autovla,mao2023gptdriver,huang2025drivegpt,rowe2025poutine}.
EMMA similarly serializes trajectories and other driving outputs in a shared
language-like space \citep{hwang2025emma}. Across these interfaces, continuous
action generation remains outside the VLM computation boundary or is
reformulated as discrete token prediction; DiffAdapterVLA instead evolves a
trajectory state within the frozen late-layer stack.

\noindent\textbf{Native Continuous Action Generation in VLAs.}
Recent robot policies and driving VLAs increasingly use unified generative
architectures in which continuous actions or trajectories evolve with visual,
linguistic, and historical context. Diffusion-transformer policies model
continuous actions through multimodal Transformer computation
\citep{liu2025rdt,reuss2024mdt,yang2025ppi,li2025uva,hou2025dita,wen2025dvla},
either placing noisy action tokens alongside scene, language, and task tokens
or jointly modeling actions with video dynamics and multimodal reasoning
\citep{yang2025ppi,li2025uva,wen2025dvla}. DITA specifically replaces a small
denoising head conditioned on fused embeddings with in-context conditioning on
raw visual tokens \citep{hou2025dita}; MindVLA-U1 similarly combines
autoregressive language generation and flow-matching trajectories under a
shared VLM backbone \citep{huang2026mindvla}.

\section{Preliminaries}
\label{sec:preliminaries}

\noindent\textbf{Frozen driving-VLM planning setup.}
We consider planning as the prediction of a future continuous trajectory from
multimodal driving conditions. Given the condition at time $t$,
\begin{equation}
    \mathbf{c}_t = \big(\mathbf{I}_t^{1:M}, \mathbf{r}_t, \mathbf{h}_t\big),
\end{equation}
where $\mathbf{I}_t^{1:M}$ denotes visual observations from $M$ cameras,
$\mathbf{r}_t$ denotes route or navigation information, and $\mathbf{h}_t$
denotes vehicle history and other available driving context, the goal is to
generate a continuous trajectory of $H$ future waypoints,
\begin{equation}
    \mathbf{x}_0 =
    [\mathbf{p}_{t+1}, \ldots, \mathbf{p}_{t+H}]
    \in \mathbb{R}^{H \times d},
\end{equation}
where $\mathbf{p}_{t+i}$ is the $i$-th future waypoint and $d$ is the waypoint
state dimension. We denote the training set by
$\mathcal{D}=\{(\mathbf{c}_t,\mathbf{x}_0)\}$.

Let $F_{\theta}$ be a pretrained driving VLM with $L$ Transformer layers. Its
condition tokens propagate through depth as
\begin{equation}
    \mathbf{H}^{\ell+1}
    = F_{\theta}^{\ell}(\mathbf{H}^{\ell}),
    \qquad \ell=0,\ldots,L-1,
\end{equation}
where $\mathbf{H}^{0}$ is encoded from visual, route, and history conditions,
and $\theta$ denotes the VLM backbone parameters. We instantiate $F_{\theta}$
with the Cosmos-Reason2-8B VLM backbone of the released Alpamayo-1.5-10B
driving VLA and keep $\theta$ frozen throughout adaptation
\citep{nvidia2025alpamayo}.

\begin{figure*}[t]
    \centering
    \includegraphics[width=0.98\textwidth]{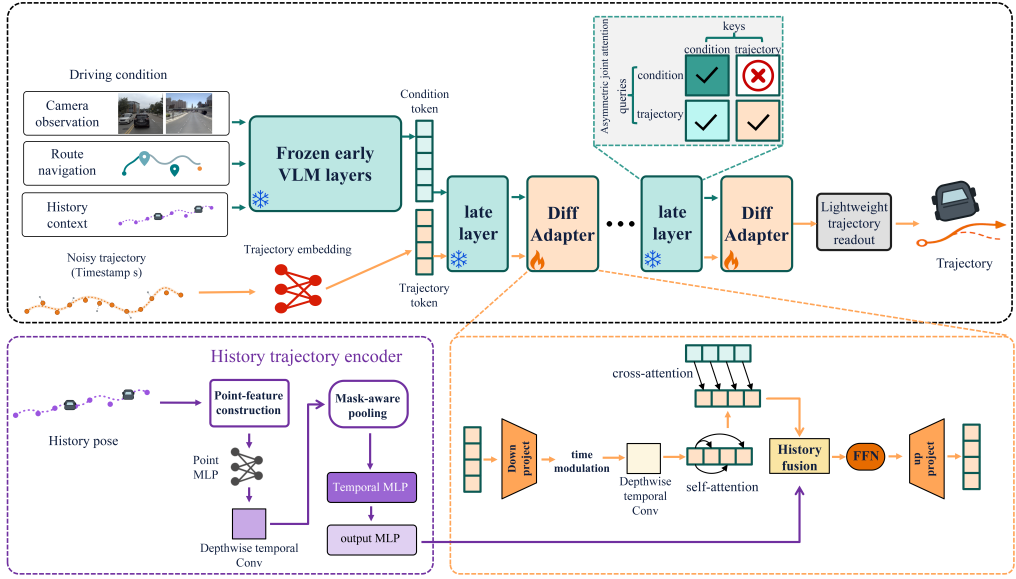}
    \caption{\textbf{Overview of DiffAdapterVLA.} Frozen early VLM layers
    encode multimodal driving conditions, while a trajectory token embedding
    maps a noisy future trajectory to trajectory tokens. Across $K$ late layers,
    interleaved DiffAdapters recursively refine and write back the trajectory
    state using the evolving condition stream and encoded motion history;
    asymmetric joint attention prevents noisy trajectory information from
    entering the condition stream. The final state is decoded to a denoising
    prediction.}
    \label{fig:architecture}
\end{figure*}

\noindent\textbf{Conditional trajectory denoising.}
We model the continuous distribution of future trajectories with conditional
diffusion. For a clean trajectory $\mathbf{x}_0$, the forward process adds
Gaussian noise at diffusion step $s\in\{1,\ldots,S\}$:
\begin{equation}
    \mathbf{x}_s
    = \sqrt{\bar{\alpha}_s}\mathbf{x}_0
    + \sqrt{1-\bar{\alpha}_s}\boldsymbol{\epsilon},
    \qquad
    \boldsymbol{\epsilon}\sim\mathcal{N}(\mathbf{0},\mathbf{I}),
\end{equation}
where $\bar{\alpha}_s$ is determined by a predefined noise schedule. A
conditional denoiser receives the noisy trajectory, diffusion step, and driving
condition to predict the added noise:
\begin{equation}
    \hat{\boldsymbol{\epsilon}}
    = \epsilon_{\phi}(\mathbf{x}_s,s;\mathbf{c}_t).
\end{equation}
It is trained with the standard noise-prediction objective
\begin{equation}
    \mathcal{L}_{\mathrm{denoise}}
    =
    \mathbb{E}_{(\mathbf{c}_t,\mathbf{x}_0)\sim\mathcal{D},\,s,\,\boldsymbol{\epsilon}}
    \left[
        \left\|
        \boldsymbol{\epsilon}
        - \epsilon_{\phi}(\mathbf{x}_s,s;\mathbf{c}_t)
        \right\|_2^2
    \right].
\end{equation}
At inference, the trajectory state is initialized from Gaussian noise and
iteratively recovered through deterministic $x_0$ projection. This process uses
the DDPM noise-prediction parameterization and follows the reduced-step
deterministic sampling principle of DDIM
\citep{ho2020ddpm,song2020ddim}. Conventional conditional trajectory
generation treats the final driving-VLM representation as a fixed condition and
performs this process in an external denoiser. The next section describes how
DiffAdapterVLA instead maintains trajectory state throughout frozen late-layer
computation.

\section{Methodology}

Figure~\ref{fig:architecture} overviews DiffAdapterVLA. Frozen early VLM
layers encode the driving condition $\mathbf{c}_t$, while a trajectory token
embedding maps $(\mathbf{x}_s,s)$ to trajectory tokens appended before the
selected $K$ late layers. Each frozen late layer updates the joint stream under
asymmetric attention, and its layer-specific DiffAdapter uses the resulting
condition tokens, diffusion step, and shared history context to update only the
trajectory tokens before writing them into the next layer. A lightweight
trajectory readout predicts the denoising target from the final state, from
which deterministic $x_0$ projection iteratively recovers the planned trajectory. We next describe the history trajectory encoder, DiffAdapter,
and condition-preserving asymmetric joint attention.

\subsection{History Trajectory Encoder}

To explicitly carry motion continuity from the observed past into future
trajectory denoising, the history trajectory encoder maps ego-motion history to
a compact context vector $\mathbf{g}_t$ that anchors denoising in the vehicle's
current motion state. Let $\mathbf{h}^{\mathrm{ego}}_t=[\mathbf{q}_{t-T_h+1},\ldots,
\mathbf{q}_t]\in\mathbb{R}^{T_h\times3}$ denote the ego-position history and
let $m_i\in\{0,1\}$ indicate whether its $i$-th position is valid. For each
position, we compute the planar displacement
$\Delta\mathbf{q}^{xy}_i=\mathbf{q}^{xy}_i-\mathbf{q}^{xy}_{i-1}$, with a
zero displacement at the first position, and form the seven-dimensional
descriptor
\begin{equation}
    \mathbf{r}_i=[\mathbf{q}_i,\Delta\mathbf{q}^{xy}_i,m_i,\rho_t],
\end{equation}
where $\rho_t$ is a valid-history statistic supplied with the input, or the
fraction of valid positions when that statistic is unavailable. This descriptor
retains absolute position, local planar motion, and observation validity.

Each $\mathbf{r}_i$ is linearly projected and normalized into the hidden space;
invalid positions are then masked. Denoting the projected representation at
waypoint $i$ by $\mathbf{u}_i$, a depthwise temporal convolution with kernel
size three produces $\tilde{\mathbf{u}}_i$ by capturing local motion variation
without mixing hidden channels:
\begin{equation}
    \tilde{\mathbf{u}}_i
    =\mathbf{u}_i+\operatorname{DWConv}_{3}(\mathbf{u})_i.
\end{equation}
We aggregate the resulting sequence by masked mean pooling,
\begin{equation}
    \bar{\mathbf{u}}_t=
    \frac{\sum_{i=1}^{T_h}m_i\tilde{\mathbf{u}}_i}
    {\max(\sum_{i=1}^{T_h}m_i,1)},
\end{equation}
so invalid positions do not contribute to the summary.

Two residual MLPs further transform the pooled motion representation:
\begin{equation}
    \mathbf{v}_t=\bar{\mathbf{u}}_t+f_{\mathrm{temp}}(\bar{\mathbf{u}}_t),
    \qquad
    \mathbf{g}_t=\mathbf{v}_t+
    f_{\mathrm{out}}\!\left(\operatorname{LN}(\mathbf{v}_t)\right).
\end{equation}
The intermediate state $\mathbf{v}_t$ is refined by a LayerNorm--SiLU MLP,
and $f_{\mathrm{out}}$ is a two-layer GELU MLP. The resulting
$\mathbf{g}_t\in\mathbb{R}^{D}$ is shared by all selected late-layer
DiffAdapters, supplying persistent history-dependent motion context to their
trajectory-state updates.

\subsection{DiffAdapter}

DiffAdapterVLA represents the noisy future trajectory
$\mathbf{x}_s\in\mathbb{R}^{H\times d}$ with trajectory tokens in the VLM
hidden space. For its $i$-th waypoint, the linear trajectory embedding adds a
future-position embedding, a trajectory token-type embedding, and a diffusion
step embedding to the projected state:
\begin{equation}
    \mathbf{z}^{0}_{s,i}=W_{\mathrm{traj}}\mathbf{x}_{s,i}
    +\mathbf{e}^{\mathrm{pos}}_i+\mathbf{e}^{\mathrm{type}}
    +\mathbf{e}^{\mathrm{time}}_s.
    \label{eq:traj-embedding}
\end{equation}
The learned projection $W_{\mathrm{traj}}$ maps each waypoint state into the
VLM hidden space, while the three additive terms encode its future position,
trajectory-token identity, and diffusion step; the last is obtained by passing
a sinusoidal timestep embedding through a SiLU MLP. After LayerNorm, the
resulting trajectory tokens are appended to the condition tokens. The first
$L-K$ frozen VLM layers propagate the multimodal condition $\mathbf{c}_t$:
\begin{equation}
    \mathbf{C}^{L-K}=F^{L-K-1}_{\theta}\circ\cdots\circ
    F^{0}_{\theta}(\mathbf{c}_t).
    \label{eq:early-condition-propagation}
\end{equation}
Here, $F^{\ell}_{\theta}$ is the $\ell$-th frozen VLM layer. These early
layers retain the VLM's original computation for integrating visual
observations, route information, and driving context before trajectory tokens
enter the network.

The appended tokens then traverse the final $K$ frozen VLM layers. Each selected
layer $\ell$ is followed by an independent DiffAdapter, with a 384-dimensional
bottleneck and eight attention heads. Let
\begin{equation}
    [\tilde{\mathbf{C}}^{\ell+1},\tilde{\mathbf{Z}}^{\ell+1}_s]
    =F^{\ell}_{\theta}([\mathbf{C}^{\ell},\mathbf{Z}^{\ell}_s])
    \label{eq:joint-late-layer}
\end{equation}
denote the joint-layer output, partitioned into condition and trajectory
slices. Tildes mark the states immediately after the frozen VLM layer.
DiffAdapter updates only the trajectory slice:
\begin{equation}
    \mathbf{C}^{\ell+1}=\tilde{\mathbf{C}}^{\ell+1},\qquad
    \mathbf{Z}^{\ell+1}_s=\mathcal{A}_{\ell}
    (\tilde{\mathbf{Z}}^{\ell+1}_s,
    \tilde{\mathbf{C}}^{\ell+1},\mathbf{g}_t,s).
    \label{eq:adapter-writeback}
\end{equation}
The operator $\mathcal{A}_{\ell}$ denotes the DiffAdapter assigned to layer
$\ell$. Its output is written into the input of the next frozen VLM layer,
whereas the condition slice follows the original VLM propagation. The following
asymmetric-attention subsection specifies the mask used for this joint
computation.

Within the adapter, the incoming trajectory state is first normalized and
projected into a bottleneck space. Because the appropriate update changes with
the noise level, a sinusoidal embedding of $s$ is passed through a SiLU MLP to
produce affine modulations for temporal mixing and self-attention, together
with residual gates for temporal mixing, self-attention, history fusion, and
the feed-forward network. The temporal and self-attention stages are then
written as
\begin{equation}
\begin{gathered}
\mathbf{u}_0=W_{\mathrm{down}}\operatorname{LN}
    (\tilde{\mathbf{Z}}^{\ell+1}_s), \\[2pt]
\mathbf{u}_1=\mathbf{u}_0+\tanh(g_{\mathrm{tmp}}(s))\odot
    \operatorname{DWConv}_{3}\!\left(
    \operatorname{AdaLN}_{\mathrm{tmp}}(\mathbf{u}_0;s)\right), \\[2pt]
\mathbf{u}_2=\mathbf{u}_1+\tanh(g_{\mathrm{self}}(s))\odot
    \operatorname{MultiHeadSelfAttention}\!\left(
    \operatorname{AdaLN}_{\mathrm{self}}(\mathbf{u}_1;s)\right).
\end{gathered}
\label{eq:adapter-temporal-self}
\end{equation}
Both $\operatorname{AdaLN}_{\mathrm{tmp}}(\cdot;s)$ and
$\operatorname{AdaLN}_{\mathrm{self}}(\cdot;s)$ use the timestep-dependent
affine parameters, while $g_{\mathrm{tmp}}(s)$ and $g_{\mathrm{self}}(s)$
scale their residual updates. Thus, the depthwise convolution mixes neighboring
future waypoints, while multi-head self-attention captures their long-range
dependencies; $\mathbf{u}_1$ and $\mathbf{u}_2$ are the resulting intermediate
trajectory states.

The adapter reads the current condition representation through cross-attention,
using trajectory states as queries and projected condition tokens as keys and
values:
\begin{equation}
    \mathbf{u}_3=\mathbf{u}_2+
    \operatorname{CrossAttn}\!\left(
    \operatorname{LN}(\mathbf{u}_2),
    W_c\operatorname{LN}(\tilde{\mathbf{C}}^{\ell+1}),
    W_c\operatorname{LN}(\tilde{\mathbf{C}}^{\ell+1})
    \right),
    \label{eq:adapter-cross-attn}
\end{equation}
The learned projection $W_c$ maps normalized condition tokens into the adapter
bottleneck space, and invalid condition positions are excluded as keys. The
projected history summary is broadcast to every trajectory token and fused with
the current state by a gated GELU MLP, followed by a gated feed-forward MLP.
The resulting bottleneck representation is up-projected and added to the
adapter input through an outer residual connection. Each layer-specific adapter
therefore combines local motion mixing, trajectory-wide coordination, condition
reading, and history-dependent refinement.

After $K$ alternating frozen-layer and DiffAdapter updates, the final trajectory
tokens are normalized by the VLM's final normalization layer and decoded by a
lightweight linear trajectory readout. Under the DDPM-noise objective, it
predicts
\begin{equation}
    \hat{\boldsymbol\epsilon}_s=
    W_{\mathrm{out}}\mathbf{Z}^{L}_s+\mathbf{b}_{\mathrm{out}}.
    \label{eq:trajectory-readout}
\end{equation}
The learned linear parameters $W_{\mathrm{out}}$ and
$\mathbf{b}_{\mathrm{out}}$ map the final trajectory state to the predicted
noise $\hat{\boldsymbol\epsilon}_s$ at step $s$. Continuous trajectory
generation is consequently realized by recursive denoising updates interleaved
with frozen late-layer computation, rather than by an independent planner
operating on the final VLM cache.

\paragraph{Denoising Inference and Condition Reuse.}

Starting from Gaussian noise, we recover the planned trajectory through five
equally spaced deterministic $x_0$-projection steps. Each step estimates a
clipped clean trajectory from the predicted noise and rescales it to the next
noise level; the complete update is given in the appendix. Since the driving
condition remains fixed throughout denoising, $\mathbf{C}^{L-K}$, its validity
mask, and position information can be cached and reused across steps. Each step
embeds the current $\mathbf{x}_s$, appends its trajectory tokens to the
condition states, and executes the late VLM layers and their layer-specific
DiffAdapters. The history encoder likewise computes $\mathbf{g}_t$ once per
sample and shares it across all denoising steps and selected late layers.

\subsection{Condition-Preserving Asymmetric Joint Attention}

The final $K$ VLM layers receive a joint sequence of condition and trajectory
tokens, whose two components have asymmetric read permissions. Let
$\mathcal{I}_{\mathrm{c}}$ and $\mathcal{I}_{\mathrm{z}}$ denote the index sets
of condition and trajectory tokens, respectively. At each late layer, we extend
the VLM's original condition mask with the joint attention mask
\begin{equation}
\mathbf{M}_{ij}=
\begin{cases}
\mathbf{M}^{\mathrm{cond}}_{ij}, & i,j\in\mathcal{I}_{\mathrm{c}},\\
-\infty, & i\in\mathcal{I}_{\mathrm{c}},\ j\in\mathcal{I}_{\mathrm{z}},\\
0, & i\in\mathcal{I}_{\mathrm{z}},\ j\in\mathcal{I}_{\mathrm{z}},\\
0, & i\in\mathcal{I}_{\mathrm{z}},\ j\in\mathcal{I}_{\mathrm{c}}\ \text{and}\ j\ \text{is valid},\\
-\infty, & i\in\mathcal{I}_{\mathrm{z}},\ j\in\mathcal{I}_{\mathrm{c}}\ \text{and}\ j\ \text{is invalid}.
\end{cases}
\label{eq:asymmetric-joint-mask}
\end{equation}
Here, $\mathbf{M}^{\mathrm{cond}}$ retains the VLM's causal attention mask on
the condition stream. Condition queries are therefore prevented from attending
to the noisy trajectory suffix, while trajectory tokens attend to all valid
condition tokens and interact bidirectionally with one another. This preserves
the condition stream's original propagation while allowing the trajectory state
to use the current-depth driving representation, other future waypoints, and
the layer-wise DiffAdapter update. The mask thus defines directed information
flow within the shared late-layer stack: driving conditions guide trajectory
denoising without noisy trajectory information entering the condition stream.

\section{Experiments}

\subsection{Experimental Setup}

\paragraph{Implementation Details.}
We build on the Alpamayo-1.5-10B and keep its Cosmos-Reason2-8B
driving VLM backbone frozen throughout training \citep{nvidia2025alpamayo}. We train only the
linear trajectory embedding, DiffAdapters, history trajectory encoder, and
lightweight trajectory readout, comprising 3.05\% of the model parameters.
DiffAdapterVLA inserts twelve layer-specific DiffAdapters into the final
$K=12$ VLM layers. We first perform supervised fine-tuning (SFT) of these modules for
80 epochs with the DDPM noise-prediction objective and 100 diffusion steps.
Starting from the SFT checkpoint, we apply DiffGRPO for a further 10 epochs to
the same trainable modules \citep{li2026recogdrive}. We use five deterministic
$x_0$-projection steps at inference; early-layer condition-hidden caching is
enabled for efficiency evaluation. All experiments are conducted on eight
NVIDIA A800 GPUs; detailed hyperparameter settings are provided in the appendix.

\paragraph{Dataset.}
We evaluate DiffAdapterVLA on NAVSIM, a planning-oriented autonomous driving
benchmark \citep{dauner2024navsim}. NAVSIM is built on OpenScene, a
resampling and curation of nuPlan driving data
\citep{caesar2021nuplan}. Its official split comprises navtrain,
with 1,192 training scenes, and navtest, with 136 evaluation scenes. Following
the NAVSIM planning protocol, the model receives the current front-camera
observation, route information, and four historical vehicle states, and
predicts eight future ego-pose waypoints at 2~Hz, corresponding to a 4-second
planning horizon. We evaluate planning quality using the official Predictive
Driver Model Score (PDMS), which jointly measures progress, safety, comfort,
and traffic-rule compliance.

\subsection{Main Results and Ablation Study}

\input{main_results_table}

\paragraph{Experiments on the NAVSIM Benchmark.}
Table~\ref{tab:navsim_main_results} reports closed-loop performance on NAVSIM
navtest. With the driving VLM frozen, DiffAdapterVLA trains only 3.05\% of
the parameters and attains 88.3 PDMS from supervised trajectory adaptation
alone, matching the LiDAR-equipped WoTE and the camera-only DP-VLA. Its RFT
model reaches 90.3 PDMS, exceeding DriveDPO by 0.3 points and DriveVLA-W0 by
0.1 points while using only the front camera. The balanced safety and progress
scores show that the trajectory adaptation preserves driving feasibility while
maintaining effective closed-loop motion.

\paragraph{Inference Efficiency.}
As shown in Table~\ref{tab:inference_speed}, with the same Cosmos-Reason2 VLM
frozen and only a downstream continuous-generation module trained,
DiffAdapterVLA improves PDMS from 71.1 to 88.3 over Alpamayo's second-stage
action expert while reducing end-to-end latency per sample from 1003.8 to
80.4~ms, a 12.5$\times$ speedup. Compared with the lightweight DiT planner
with LoRA-adapted VLM representations, DiffAdapterVLA operates in a comparable
latency regime (80.4 versus 67.6~ms) while improving PDMS by 6.4 points.
We further compare representative systems based on an external diffusion
planner, an external action expert, and autoregressive action tokens:
ReCogDrive~\citep{li2026recogdrive}, DriveVLA-W0~\citep{li2026drivevlaw0},
and OneVL~\citep{lu2026onevl}, respectively. Without sacrificing
PDMS, DiffAdapterVLA reduces end-to-end latency per sample by 9.2$\times$,
14.8$\times$, and 15.7$\times$, respectively, while improving PDMS by 1.8,
1.1, and 0.8 points. These cross-paradigm comparisons show that incorporating
trajectory refinement into VLM late-layer computation provides a more favorable
efficiency--performance trade-off for continuous planning: high-quality
trajectories need not be generated solely by an independent module outside the
VLM.

\paragraph{Ablation Study on DiffAdapterVLA.}
Table~\ref{tab:component_ablations} presents ablations of DiffAdapterVLA's core
components and its asymmetric joint attention mechanism. Omitting
both the DiffAdapter and history trajectory encoder reduces PDMS to 58.2,
highlighting the importance of explicit trajectory-state updates for continuous
planning. With the DiffAdapter retained, removing the history encoder and using
bidirectional interaction yield 85.3 and 85.6 PDMS, respectively, both below
the full model's 86.9. The full design also attains the highest EP of 80.5,
indicating that motion history and condition-preserving interaction jointly
support trajectory generation.

\paragraph{Qualitative Analysis.}
Figure~\ref{fig:qualitative_navsim} compares trajectories from DiffAdapterVLA,
OneVL, and ReCogDrive. Across the two NAVSIM navtest scenes, DiffAdapterVLA
more closely follows the ground-truth trajectory, particularly through the
curved maneuver in the roundabout scene. Additional visualizations are provided
in the appendix.

\input{evidence_float}

\paragraph{Does Trajectory Generation Benefit from Backbone Computation?}
Table~\ref{tab:condition_integration} examines three ways of using driving
conditions for trajectory generation: generating from a final condition only,
reading late-layer conditions while keeping trajectory updates outside the VLM,
and jointly evolving trajectory tokens with conditions in late-layer computation.
Replacing the final condition with layer-wise late-layer conditions raises PDMS
from 80.2 to 85.2; incorporating trajectory tokens into the corresponding
backbone computation further raises it to 86.9. Thus, continuously using
driving conditions from different depths is more effective than decoding a
trajectory once from a completed final condition, and joint forward computation
of trajectory tokens and conditions yields an additional gain. To examine how
this process unfolds at inference, Figure~\ref{fig:layerwise_conditioning}
records the trajectory-state update and trajectory-to-condition cross-attention
residual of each DiffAdapter at every denoising step: panel~(a) reports the
base-10 logarithm of update RMS normalized by incoming trajectory-state RMS,
and panel~(b) reports residual RMS. Both signals vary substantially across depth
and denoising time. At the noisy step $t{=}80$, adapters 0 and 1 produce
relative trajectory updates of about 68\% and 70\%, respectively; as denoising
approaches $t{=}0$, the updates of adapters 7--10 grow from about 15--19\% to
27--32\%. Condition residuals persist throughout the late-layer stack, range
from 0.15 to 0.78, and are strongest at intermediate adapters 4--5 during
mid-stage denoising. The full model therefore does not decode a trajectory once
from a completed condition: driving conditions at different depths contribute to
trajectory refinement at different stages of denoising. Placing trajectory tokens
in backbone late-layer computation, where they evolve jointly with condition
information, provides direct support for backbone-native continuous trajectory
planning.

\section{Conclusion}

We present DiffAdapterVLA, a lightweight adaptation framework that brings
continuous trajectory planning into the backbone computation of a driving VLM.
Existing VLAs typically separate understanding from planning: the VLM produces
a final condition before a subsequent module generates the trajectory, treating
pretrained driving priors as inputs to planning rather than participants in
trajectory formation. DiffAdapterVLA instead introduces explicit trajectory
state into selected late layers, where it evolves with depth-wise driving
conditions throughout denoising and turns representation formation into
layer-wise trajectory refinement. With lightweight DiffAdapters and directed
information exchange, the framework trains only a small set of parameters and
achieves high-quality, low-latency continuous planning without an independent
trajectory planner. Results on NAVSIM validate its planning performance and
efficiency, while showing the benefit of trajectory state participating in
backbone condition computation rather than reading the final condition once.
DiffAdapterVLA therefore provides a new path between driving understanding and
continuous planning, enabling existing driving priors to directly shape future
motion through parameter-efficient adaptation.

\section*{AI Use Statement}
In this work, we used generative AI tools for related-work retrieval and
discovery, research ideation, and implementation assistance, including coding
support. We have not used generative AI tools for generating synthetic datasets
or proving mathematical claims, and the remaining required disclosure tasks are
not applicable to this work. Additionally, we used generative AI tools to edit
the manuscript for grammar, wording, and clarity. All AI-assisted work was
reviewed and verified by the authors. The authors take responsibility for the
final content of this work, including text, claims, and artifacts produced with
the aid of generative AI.
\bibliographystyle{references}
\bibliography{references}

\clearpage
\appendix
\section{APPENDIX}
\label{app:overview}

This appendix provides the full implementation details and evaluation protocol,
additional results, the limitations and future directions of the approach, and further qualitative
trajectory comparisons.

\section{Full Implementation Details}
\label{app:implementation}

\paragraph{Evaluation Metric.}
We use the official Predictive Driver Model Score (PDMS) from NAVSIM to
evaluate closed-loop planning. The PDM simulator rolls each predicted
trajectory out at 10~Hz and evaluates no-at-fault collision (NC), drivable-area
compliance (DAC), time-to-collision (TTC), comfort (C), and ego progress (EP).
Let $n,d,\tau,c,e\in[0,1]$ denote these quantities. NC is determined by the
official collision-responsibility rule, while DAC requires the ego footprint to
remain in the drivable area throughout the rollout:
\begin{equation}
d=\prod_{t=1}^{T}\mathbf{1}\!\left[\mathcal{B}_{t}\subseteq\mathcal{D}\right],
\end{equation}
where $\mathcal{B}_{t}$ is the ego footprint and $\mathcal{D}$ is the
drivable area. TTC checks for a collision under a short-horizon extrapolation
of the ego motion at each rollout state:
\begin{equation}
\tau=\prod_{t=1}^{T}\mathbf{1}\!\left[\mathrm{TTC}_{t}>0\right].
\end{equation}
Comfort jointly constrains longitudinal and lateral acceleration, total and
longitudinal jerk, yaw acceleration, and yaw rate. With $\mathcal{Q}$ denoting
these six quantities and $[l_q,u_q]$ their official bounds, it is computed as
\begin{equation}
c=\prod_{q\in\mathcal{Q}}\prod_{t=1}^{T}
\mathbf{1}\!\left[l_q\leq q_t\leq u_q\right].
\end{equation}
The progress term is safety-gated by $g=nd$ and normalized against the largest
gated progress among the prediction and the PDM reference trajectory evaluated
in the same scenario:
\begin{equation}
e=
\begin{cases}
\dfrac{g[\Delta p]_+}{\max_j\left(g_j[\Delta p_j]_+\right)}, &
\max_j\left(g_j[\Delta p_j]_+\right)>5\,\mathrm{m},\\[3pt]
\mathbf{1}[g>0], & \text{otherwise},
\end{cases}
\end{equation}
where $\Delta p$ is the forward centerline progress and $j$ indexes the two
proposals. PDMS combines the resulting terms as
\begin{equation}
\mathrm{PDMS}=nd\cdot\frac{5e+5\tau+2c}{12}.
\end{equation}
All values in the paper are averaged over NAVSIM navtest and reported on a
percentage scale. The official scorer also computes driving-direction
compliance, whose weight is zero in the NAVSIM configuration used here.

\paragraph{Training Protocol for Main Results.}
The SFT and RFT models in Table~\ref{tab:navsim_main_results} share the same
architecture, input protocol, and trainable parameter set. They receive the
current front-camera observation, route information, and four historical ego
states, and predict eight future ego poses. Following the configuration in the
main text, the twelve DiffAdapters operate on the final $K=12$ VLM layers, each
with a 384-dimensional bottleneck and eight attention heads. SFT uses the DDPM
noise-prediction objective with 100 diffusion steps and an 80-epoch training
schedule. It is trained on eight NVIDIA A800 GPUs with bfloat16 precision and
ZeRO-2 optimization, using a per-device batch size of 16, a global batch size of
128, a learning rate of $10^{-5}$, 500 warm-up steps, and cosine decay to
$10^{-6}$. Future poses are normalized using the training-data pose range and
clipped during sampling. Starting from SFT, RFT applies DiffGRPO with official
PDMS rewards. Eight trajectories are sampled per driving condition for the
group-relative update, together with a teacher-chain behavioral-cloning anchor
and an SFT reference regularizer weighted by $0.1$ and $0.05$, respectively.
RFT uses a 10-epoch schedule on eight NVIDIA A800 GPUs, with a per-device batch
size of 8, global batch size of 64, learning rate $10^{-5}$ with cosine decay to
$10^{-6}$, weight decay $10^{-4}$, Adam coefficients $(0.9,0.95)$, and gradient
clipping at $1.0$.

\paragraph{Deterministic Denoising Process.}
Given trajectory state $\mathbf{x}_s$ and predicted noise
$\hat{\boldsymbol{\epsilon}}_s$ at step $s$, we estimate the normalized clean
trajectory as
\begin{equation}
    \hat{\mathbf{x}}_0^{(s)}
    =
    \operatorname{clip}\!\left(
    \frac{\mathbf{x}_s-\sqrt{1-\bar{\alpha}_s}\,
    \hat{\boldsymbol{\epsilon}}_s}
    {\sqrt{\bar{\alpha}_s}},
    -1,1
    \right).
\end{equation}
For the next scheduled step $s'<s$, the trajectory state is updated by
\begin{equation}
    \mathbf{x}_{s'}=
    \sqrt{\bar{\alpha}_{s'}}\,\hat{\mathbf{x}}_0^{(s)}.
\end{equation}
The final step returns its clean-trajectory estimate. We use five equally
spaced steps in the main experiments.

\paragraph{Inference Protocol for Table~\ref{tab:inference_speed}.}
Table~\ref{tab:inference_speed} compares trajectory-generation paradigms under
a shared eight-GPU NAVSIM protocol. Each system processes the same front-camera
planning input with a batch size of 8; measurements use fixed input shapes after
warm-up and report a single output trajectory per input. \emph{Model
ms/sample} measures model forward computation and trajectory generation,
whereas \emph{E2E ms/sample} additionally includes input preparation,
trajectory post-processing, and the evaluation interface. \emph{E2E batch
time} is the mean wall-clock time for one complete batch. DiffAdapterVLA uses
five deterministic $x_0$-projection steps and reuses condition hidden states computed by
the frozen early VLM layers across denoising steps.

\paragraph{Protocols for Component and Interface Studies.}
All variants in Table~\ref{tab:component_ablations} are trained on navtrain for
50 epochs with the same inputs, DDPM objective, optimizer configuration, and
five-step deterministic $x_0$-projection inference. The full model enables asymmetric joint attention,
the history trajectory encoder, and DiffAdapter. The history ablation removes
the global motion context obtained from historical ego states; the attention
ablation replaces directed interaction with bidirectional interaction; and the
DiffAdapter ablation removes layer-wise trajectory-state updates. Table~\ref{tab:condition_integration}
keeps the backbone, adapter capacity, training data, and sampling schedule
fixed while varying the trajectory--VLM interface. The three settings provide,
respectively, only a final condition, layer-wise late-layer conditions while
trajectory states remain outside VLM layers, and joint late-layer computation
of trajectory and condition tokens.

\paragraph{Layer-wise Dynamic Analysis.}
For Figure~\ref{fig:layerwise_conditioning}, we record the trajectory-state
update and trajectory-to-condition cross-attention residual of every
DiffAdapter at each denoising step. Panel~(a) reports the base-10 logarithm of the
update RMS normalized by the incoming trajectory-state RMS,
\begin{equation}
\log_{10}\!\left(
\frac{\operatorname{RMS}(\Delta\mathbf{Z}_{l,s})}
{\operatorname{RMS}(\mathbf{Z}^{\mathrm{in}}_{l,s})}
\right),
\end{equation}
where $\Delta\mathbf{Z}_{l,s}$ is the update from adapter $l$ at denoising step
$s$. Panel~(b) reports the RMS of its trajectory-to-condition cross-attention
residual $\mathbf{R}^{\mathrm{cross}}_{l,s}$:
\begin{equation}
\operatorname{RMS}\!\left(\mathbf{R}^{\mathrm{cross}}_{l,s}\right).
\end{equation}
These activation statistics characterize the depth- and timestep-dependent
roles of trajectory refinement and condition access during inference.

\section{More Results}
\label{app:more_results}

\paragraph{Late-layer Span for Backbone-Native Planning.}
Table~\ref{tab:late_layer_count} examines how broadly trajectory states should
participate in the frozen VLM's late-layer computation. This directly probes the
structural premise of DiffAdapterVLA: continuous planning should neither reduce
to one-shot decoding from only a few final representations nor enter backbone
computation before multimodal driving conditions are formed. Expanding the
selected span from $K=4$ to $K=12$ raises PDMS from 73.0 to 86.9, with EP
increasing from 67.0 to 80.5, showing that recursive access to conditions across
multiple late layers substantially improves effective closed-loop progress and
overall planning quality. Extending the span further to $K=16$ and $K=20$ lowers
PDMS to 82.1 and 81.1, respectively. Although both retain high NC and TTC,
their DAC and EP decline. The strongest planning interval therefore lies in the
backbone's later portion: a sufficiently broad late-layer stack supplies
trajectory refinement with progressively structured driving conditions, while
leaving earlier layers to form those conditions preserves the intended division
of computation.

\begin{table}[H]
    \centering
    \small
    \setlength{\tabcolsep}{3.3pt}
    \localtablecaption{Effect of the number of selected late VLM layers. $K$ denotes the number of final VLM layers equipped with
    DiffAdapters.}
    \label{tab:late_layer_count}
    \begin{tabular}{@{}c c c c c c c c@{}}
        \toprule
        $K$ & Trainable (\%) & NC$\uparrow$ & DAC$\uparrow$ & TTC$\uparrow$ & C$\uparrow$ &
        EP$\uparrow$ & PDMS$\uparrow$ \\
        \midrule
        4  & 2.03 & 95.4 & 88.1 & 85.2 & 100.0 & 67.0 & 73.0 \\
        8  & 2.54 & 98.4 & 93.7 & 94.9 & 100.0 & 75.7 & 84.1 \\
        12 & 3.05 & 98.1 & 95.7 & 94.8 & 100.0 & 80.5 & 86.9 \\
        16 & 3.56 & 98.0 & 91.5 & 94.9 & 100.0 & 74.2 & 82.1 \\
        20 & 4.06 & 98.8 & 91.0 & 95.7 & 100.0 & 71.1 & 81.1 \\
        \bottomrule
    \end{tabular}
\end{table}

\paragraph{Quality--Efficiency Trade-off across Denoising Steps.}
Table~\ref{tab:ddim_step_sweep} varies only the number of deterministic
$x_0$-projection steps for the same SFT checkpoint. A single step does not recover a usable
trajectory, whereas planning quality improves steadily from two to five steps.
The gain then nearly saturates: increasing from five to ten steps changes PDMS
from 87.00 to 87.07, while E2E latency rises from 80.8 to 126.8~ms per sample.
Five steps therefore provide the operating point used in the main experiments,
retaining near-saturated planning quality without the additional iterative cost
of ten-step sampling.

\begin{table}[H]
    \centering
    \small
    \setlength{\tabcolsep}{4.0pt}
    \localtablecaption{Quality--efficiency trade-off across deterministic denoising steps; timing follows the Table~\ref{tab:inference_speed} protocol.}
    \label{tab:ddim_step_sweep}
    \begin{tabular}{@{}c c c c c@{}}
        \toprule
        Denoising steps & PDMS$\uparrow$ & E2E ms/sample$\downarrow$ &
        E2E batch time (s)$\downarrow$ & E2E samples/s$\uparrow$ \\
        \midrule
        1  & 1.92 & 74.1  & 0.593 & 101.8 \\
        2  & 82.63 & 68.5  & 0.548 & 112.0 \\
        3  & 84.32 & 72.1  & 0.577 & 106.6 \\
        5  & 87.00 & 80.8  & 0.646 & 98.0 \\
        10 & 87.07 & 126.8 & 1.014 & 62.2 \\
        \bottomrule
    \end{tabular}
\end{table}

\paragraph{Extended Predictive Driver Model Score.}
Table~\ref{tab:navsim_epdms_results} further compares planning performance under
the NAVSIM v2 extended evaluation protocol. In addition to the standard PDMS
metrics, the protocol reports Driving Direction Compliance (DDC), Traffic Light
Compliance (TLC), Lane Keeping (LK), History Comfort (HC), and Extended Comfort
(EC). DiffAdapterVLA achieves an EPDMS of 86.3, outperforming DriveVLA-W0
and ReCogDrive by 0.2 and 2.7 points, respectively. Beyond the aggregate score,
the model maintains stable driving progress and safety-related performance,
with 88.2 EP, 97.7 TTC, 99.1 DDC, and 99.8 TLC. These results show that the
model preserves effective progress while satisfying collision-risk,
driving-direction, and traffic-light constraints. To accommodate the
cross-frame motion-consistency measure in the extended evaluation, we apply
path-preserving temporal calibration to the model output. This operation
regularizes only the temporal parameterization along the existing path without
changing the predicted spatial polyline or endpoint, thereby retaining the
model's planned driving direction and spatial path. The overall results indicate
that DiffAdapterVLA benefits from a balanced combination of driving
progress, safety, and rule compliance rather than a single dominant metric.

\begin{table}[H]
    \centering
    \footnotesize
    \setlength{\tabcolsep}{3.5pt}
    \localtablecaption{Comparison on NAVSIM v2 with extended planning metrics.}
    \label{tab:navsim_epdms_results}
    \resizebox{\columnwidth}{!}{%
    \begin{tabular}{@{}l c c c c c c c c c c@{}}
        \toprule
        Method & NC$\uparrow$ & DAC$\uparrow$ & DDC$\uparrow$ & TLC$\uparrow$ &
        TTC$\uparrow$ & EP$\uparrow$ & LK$\uparrow$ & HC$\uparrow$ & EC$\uparrow$ & EPDMS$\uparrow$ \\
        \midrule
        TransFuser~\citep{chitta2022transfuser} & 97.7 & 92.8 & 98.3 & 99.9 & 92.8 & 79.2 & 67.6 & 100.0 & 95.3 & 77.8 \\
        ReCogDrive~\citep{li2026recogdrive} & 98.3 & 95.2 & 99.5 & 99.8 & 97.5 & 87.1 & 96.6 & 98.3 & 86.5 & 83.6 \\
        Hydra-MDP++~\citep{li2025hydramdpp} & 98.8 & 97.8 & 99.1 & 100.0 & 95.3 & 84.0 & 70.1 & 100.0 & 96.8 & 84.1 \\
        DiffusionDrive~\citep{liao2025diffusiondrive} & 98.2 & 95.9 & 99.4 & 99.8 & 97.3 & 87.5 & 96.8 & 98.3 & 87.7 & 84.5 \\
        Curious-VLA~\citep{chen2026curiousvla} & 98.4 & 96.9 & 99.2 & 99.8 & 97.9 & 88.5 & 96.9 & 98.1 & 81.5 & 85.3 \\
        DiffusionDriveV2~\citep{zou2025diffusiondrivev2} & 97.7 & 96.6 & 99.2 & 99.8 & 97.2 & 88.9 & 96.0 & 97.8 & 91.0 & 85.5 \\
        DriveVLA-W0~\citep{li2026drivevlaw0} & 98.5 & 99.1 & 98.0 & 99.7 & 98.1 & 86.4 & 93.2 & 97.9 & 58.9 & 86.1 \\
        \midrule
        DiffAdapterVLA (ours) & 98.4 & 98.0 & 99.1 & 99.8 & 97.7 & 88.2 & 90.3 & 97.8 & 61.4 & 86.3 \\
        \bottomrule
    \end{tabular}
    }
\end{table}

\begin{figure}[H]
    \centering
    \includegraphics[width=0.98\linewidth]{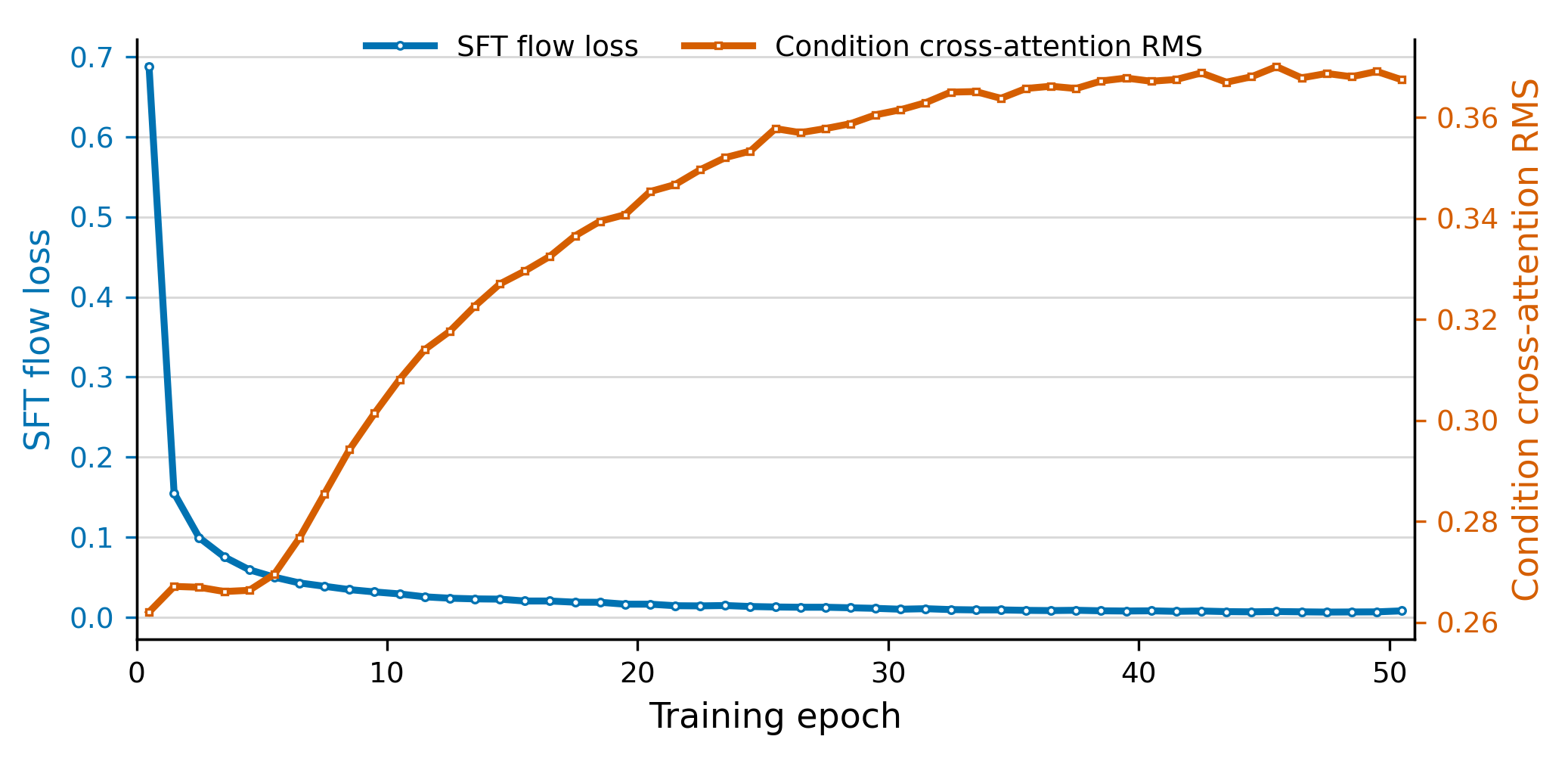}
    \localfigurecaption{\textbf{SFT training dynamics.} Epoch-wise mean SFT
    flow loss and condition cross-attention RMS. The
    two curves use separate vertical axes.}
    \label{fig:sft_condition_dynamics}
\end{figure}

\paragraph{Condition Interaction during Training.}
Figure~\ref{fig:sft_condition_dynamics} tracks the epoch-wise mean SFT flow
loss and the RMS of the trajectory-to-condition cross-attention output. The
flow loss falls rapidly at the beginning of training and then gradually
converges, while the condition-interaction RMS rises from about 0.28 to
0.36--0.37 and remains stable in later epochs. Thus, condition interaction is
not confined to an early transient; it continues to participate in trajectory
refinement at a stable scale after the denoising loss has converged.

\section{Limitations and Future Work}
\label{app:limitations}

DiffAdapterVLA writes continuous trajectory states into the late-layer
computation of a driving VLM, and its planning behavior remains bounded by the
condition information available to the backbone. The current conditions combine
visual observations, route and language prompts, and motion history; how richer
sensor, map, vehicle-to-infrastructure, or long-horizon interaction information
can participate in trajectory refinement while preserving the condition stream
is an open question. Moreover, the explicit state represents only the ego
vehicle's continuous future motion, while multimodal futures and uncertainty of
other traffic participants are primarily implicit in the condition
representation. A natural direction is to jointly maintain interactive
environment states and ego trajectories within backbone computation, enabling
planning to reason explicitly about multi-agent prediction, long-horizon
decisions, and uncertainty. Finally, NAVSIM provides a rigorous planning-oriented
closed-loop evaluation; broader road distributions, sensor degradation, and
end-to-end system latency remain important settings for assessing real-world
deployment.

\section{Additional Qualitative Results}
\label{app:additional_qualitative}

The following examples are selected from the remaining top-ranked contrastive
cases after excluding the two scenes shown in the main paper. Each row compares
OneVL, ReCogDrive, and DiffAdapterVLA against the same ground-truth trajectory.

\begin{figure*}[p]
    \centering
    \includegraphics[height=0.84\textheight,keepaspectratio]{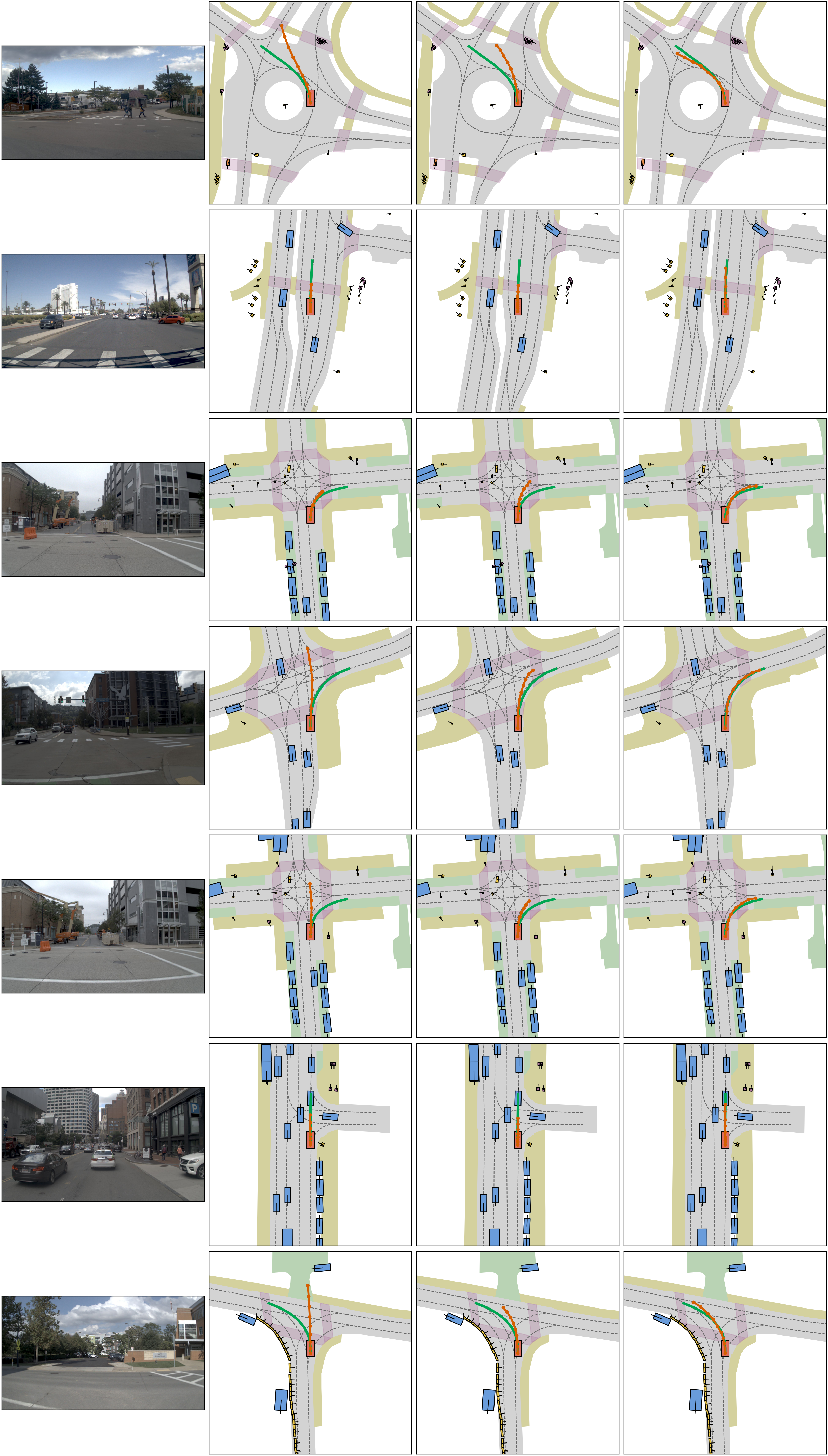}
\end{figure*}

\begin{figure*}[p]
    \centering
    \includegraphics[height=0.84\textheight,keepaspectratio]{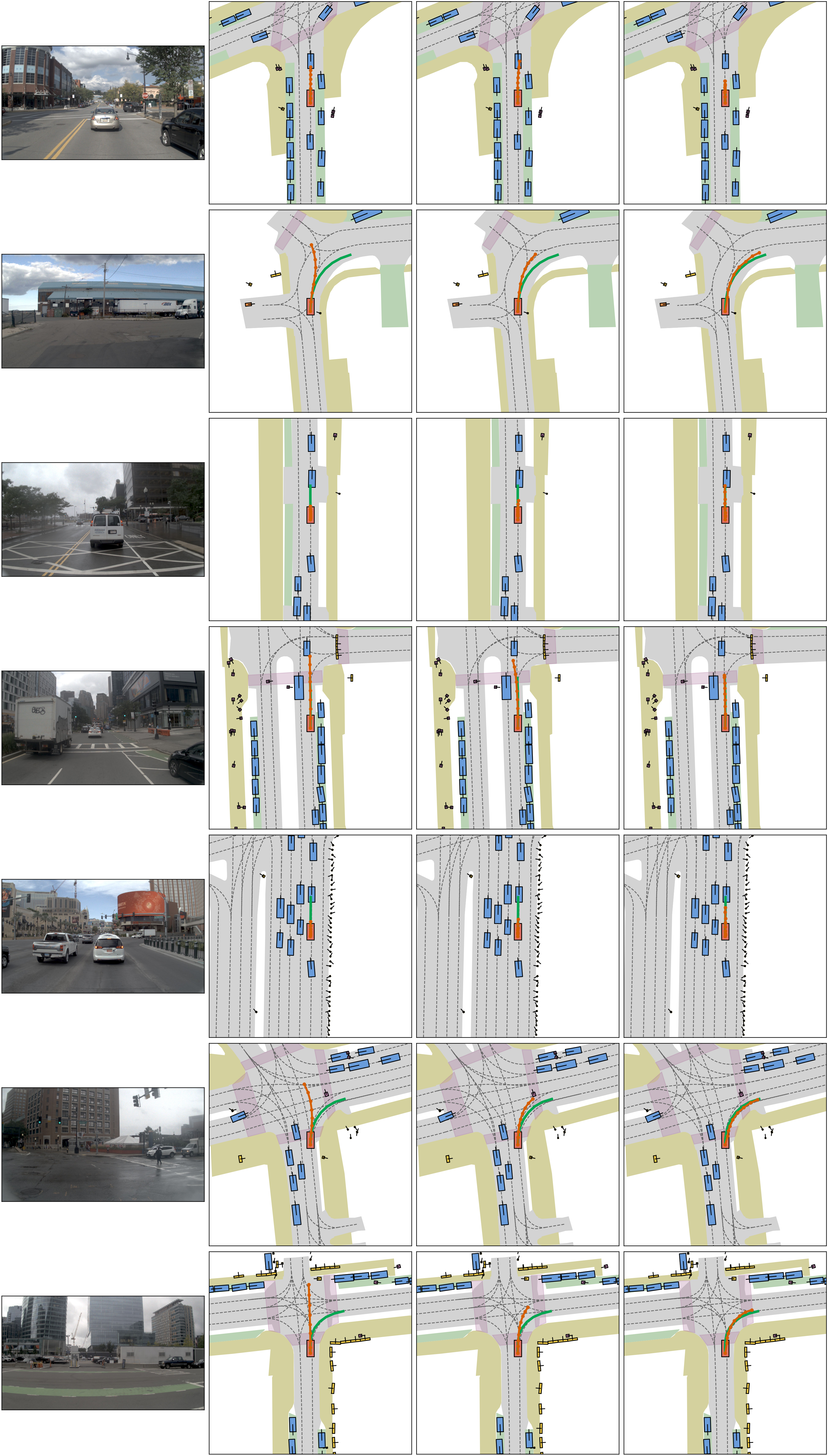}
\end{figure*}

\begin{figure*}[p]
    \centering
    \includegraphics[height=0.84\textheight,keepaspectratio]{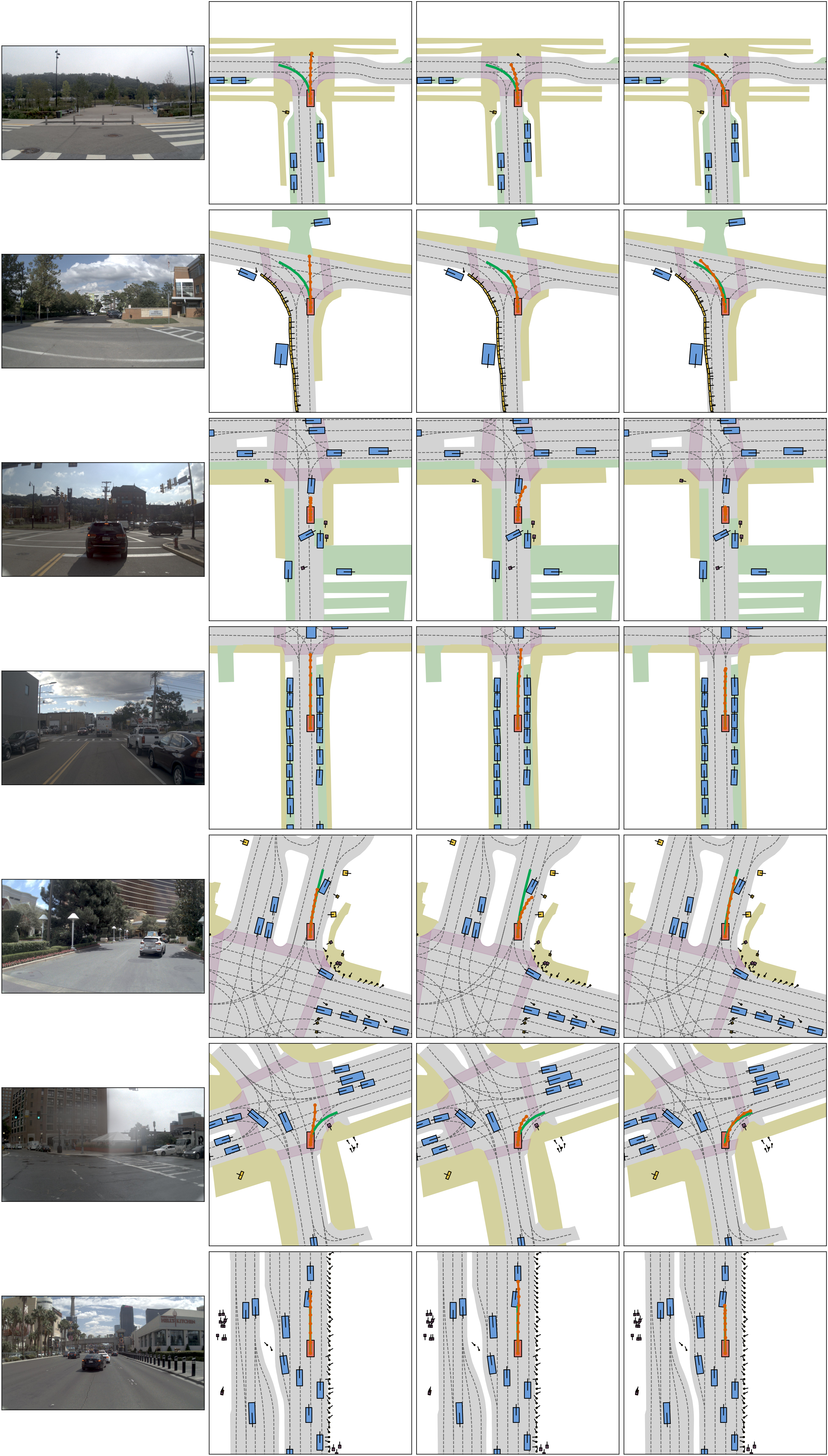}
\end{figure*}

\end{document}

%% file: math_commands.tex
\usepackage{amsmath,amsfonts,bm}

\def\eqref#1{equation~\ref{#1}}

\def\1{\bm{1}}

\DeclareMathAlphabet{\mathsfit}{\encodingdefault}{\sfdefault}{m}{sl}
\SetMathAlphabet{\mathsfit}{bold}{\encodingdefault}{\sfdefault}{bx}{n}



%% file: main_results_table.tex
\begin{table*}[t]
    \centering
    \small
    \setlength{\tabcolsep}{5pt}
    \localtablecaption{NAVSIM navtest performance comparison measured by PDMS.}
    \label{tab:navsim_main_results}
    \resizebox{\textwidth}{!}{%
    \begin{tabular}{@{}l c c c c c c c c c@{}}
        \toprule
        Method & Ref. & Image & LiDAR & NC$\uparrow$ & DAC$\uparrow$ &
        TTC$\uparrow$ & C$\uparrow$ & EP$\uparrow$ & PDMS$\uparrow$ \\
        \midrule
        Human & -- & -- & -- & 100.0 & 100.0 & 100.0 & 99.9 & 87.5 & 94.8 \\
        \midrule
        DrivingGPT~\citep{chen2025drivinggpt} & ICCV'25 & \checkmark &  & 98.9 & 90.7 & 94.9 & 95.6 & 79.7 & 82.4 \\
        UniAD~\citep{hu2023uniad} & CVPR'23 & \checkmark & \checkmark & 97.8 & 91.9 & 92.9 & 100.0 & 78.8 & 83.4 \\
        TransFuser~\citep{chitta2022transfuser} & TPAMI'23 & \checkmark & \checkmark & 97.7 & 92.8 & 92.8 & 100.0 & 79.2 & 84.0 \\
        LAW~\citep{li2025law} & ICLR'25 & \checkmark &  & 96.4 & 95.4 & 88.7 & 99.9 & 81.7 & 84.6 \\
        Epona~\citep{zhang2025epona} & ICCV'25 & \checkmark &  & 97.9 & 95.1 & 93.8 & 99.9 & 80.4 & 86.2 \\
        ReCogDrive~\citep{li2026recogdrive} & ICLR'26 & \checkmark &  & 98.1 & 94.7 & 94.2 & 100.0 & 80.9 & 86.5 \\
        DiffusionDrive~\citep{liao2025diffusiondrive} & CVPR'25 & \checkmark & \checkmark & 98.2 & 96.2 & 94.7 & 100.0 & 82.2 & 88.1 \\
        WoTE~\citep{li2025wote} & ICCV'25 & \checkmark & \checkmark & 98.5 & 96.8 & 94.9 & 99.9 & 81.9 & 88.3 \\
        DP-VLA~\citep{liang2026dipole} & ICLR'26 & \checkmark &  & 98.0 & 97.0 & 94.3 & 100.0 & 82.5 & 88.3 \\
        OneVL~\citep{lu2026onevl} & arXiv'26 & \checkmark &  & -- & -- & -- & -- & -- & 88.8 \\
        AutoVLA~\citep{zhou2025autovla} & NeurIPS'25 & \checkmark &  & 98.4 & 95.6 & 98.0 & 99.9 & 81.9 & 89.1 \\
        BrainWAM~\citep{zhan2026brainwam} & arXiv'26 & \checkmark &  & 98.1 & 97.5 & 94.9 & 100.0 & 83.8 & 89.5 \\
        CoWorld-VLA~\citep{huang2026coworldvla} & arXiv'26 & \checkmark &  & 99.2 & 96.8 & 96.6 & 100.0 & 83.6 & 89.8 \\
        DriveDPO~\citep{shang2025drivedpo} & NeurIPS'25 & \checkmark & \checkmark & 98.5 & 98.1 & 94.8 & 99.9 & 84.3 & 90.0 \\
        DriveVLA-W0~\citep{li2026drivevlaw0} & ICLR'26 & \checkmark &  & 98.7 & 99.1 & 95.3 & 99.3 & 83.3 & 90.2 \\
        \midrule
        DiffAdapterVLA-SFT (ours) & -- & \checkmark &  & 98.7 & 96.7 & 95.9 & 100.0 & 80.9 & 88.3 \\
        DiffAdapterVLA-RFT (ours) & -- & \checkmark &  & 98.8 & 98.1 & 96.7 & 100.0 & 83.0 & 90.3 \\
        \bottomrule
    \end{tabular}
    }
\end{table*}

%% file: evidence_float.tex
\begin{table*}[t]
    \centering
    \footnotesize
    \setlength{\tabcolsep}{5pt}
    \localtablecaption{Inference speed on NAVSIM under a shared 8-GPU protocol.
    Model and E2E latency denote mean milliseconds per sample. E2E batch time
    denotes mean end-to-end latency per batch.}
    \label{tab:inference_speed}
    \resizebox{\textwidth}{!}{%
    \begin{tabular}{@{}c c c c c c@{}}
        \toprule
        Model & Trajectory generation & PDMS$\uparrow$ & Model ms/sample$\downarrow$ &
        E2E ms/sample$\downarrow$ & E2E batch time (s)$\downarrow$ \\
        \midrule
        Cosmos-Reason2 VLM + DiT (no LoRA) & External DiT planner & 76.4 & 40.2 & 67.5 & 0.540 \\
        Alpamayo 1.5 (stage-2 training) & External action expert & 71.1 & 1003.0 & 1003.8 & 8.030 \\
        Cosmos-Reason2 VLM + DiT (LoRA) & External DiT planner & 81.9 & 43.6 & 67.6 & 0.541 \\
        ReCogDrive Large-IL & External diffusion planner & 86.5 & 722.5 & 739.1 & 5.913 \\
        DriveVLA-W0 (flow) & External action expert & 87.2 & 1186.8 & 1186.8 & 9.463 \\
        OneVL AR Answer & Autoregressive action tokens & 87.5 & 1218.3 & 1258.8 & 10.112 \\
        \midrule
        DiffAdapterVLA ($K{=}12$) & Native diffusion adapters & 88.3 & 77.4 & 80.4 & 0.643 \\
        \bottomrule
    \end{tabular}
    }

    \vspace{5pt}
    \begin{minipage}[t]{0.49\textwidth}
        \centering
        \scriptsize
        \setlength{\tabcolsep}{1.5pt}
        \localtablecaption{Ablation study on NAVSIM navtest (50 training epochs).}
        \label{tab:component_ablations}
        \begin{tabular}{@{}c c c c c c c@{}}
            \toprule
            \multicolumn{3}{c}{Components} & \multicolumn{4}{c}{NAVSIM metrics} \\
            \cmidrule(lr){1-3}\cmidrule(lr){4-7}
            \shortstack{Asymmetric\\joint attention} &
            \shortstack{History\\trajectory encoder} &
            \shortstack{Diff\\Adapter} & \raisebox{0.7ex}{NC} &
            \raisebox{0.7ex}{TTC} & \raisebox{0.7ex}{EP} &
            \raisebox{0.7ex}{PDMS$\uparrow$} \\
            \midrule
            \checkmark & $\times$ & $\times$ & 86.1 & 74.9 & 51.9 & 58.2 \\
            \checkmark & $\times$ & \checkmark & 98.0 & 94.2 & 78.0 & 85.3 \\
            $\times$ & \checkmark & \checkmark & 98.5 & 96.0 & 77.8 & 85.6 \\
            \checkmark & \checkmark & \checkmark & 98.1 & 94.8 & 80.5 & 86.9 \\
            \bottomrule
        \end{tabular}
    \end{minipage}\hfill
    \begin{minipage}[t]{0.49\textwidth}
        \centering
        \vspace{4pt}
        \scriptsize
        \setlength{\tabcolsep}{1.5pt}
        \renewcommand{\arraystretch}{1.40}
        \localtablecaption{Condition access for trajectory generation.}
        \label{tab:condition_integration}
        \vspace{1pt}
        \begin{tabular*}{\linewidth}{@{}c@{\extracolsep{\fill}}c c c c c c@{}}
            \toprule
            {\fontsize{7.5}{8.5}\selectfont\shortstack{Planning\\[-2pt]interface}} &
            \smash{\raisebox{0.4ex}{NC}} &
            \smash{\raisebox{0.4ex}{DAC}} &
            \smash{\raisebox{0.4ex}{TTC}} &
            \smash{\raisebox{0.4ex}{C}} &
            \smash{\raisebox{0.4ex}{EP}} &
            \smash{\raisebox{0.4ex}{PDMS$\uparrow$}} \\
            \midrule
            \shortstack{Final-cache\\[-2pt]condition only} & \smash{\raisebox{0.4ex}{95.7}} &
            \smash{\raisebox{0.4ex}{90.9}} & \smash{\raisebox{0.4ex}{89.7}} &
            \smash{\raisebox{0.4ex}{100.0}} & \smash{\raisebox{0.4ex}{76.2}} & \smash{\raisebox{0.4ex}{80.2}} \\
            \shortstack{Late-layer conditions\\[-2pt]trajectory outside VLM} & \smash{\raisebox{0.4ex}{97.6}} &
            \smash{\raisebox{0.4ex}{94.5}} & \smash{\raisebox{0.4ex}{93.3}} &
            \smash{\raisebox{0.4ex}{100.0}} & \smash{\raisebox{0.4ex}{79.3}} & \smash{\raisebox{0.4ex}{85.2}} \\
            \shortstack{Late-layer conditions\\[-2pt]trajectory inside VLM} & \smash{\raisebox{0.4ex}{98.1}} &
            \smash{\raisebox{0.4ex}{95.7}} & \smash{\raisebox{0.4ex}{94.8}} &
            \smash{\raisebox{0.4ex}{100.0}} & \smash{\raisebox{0.4ex}{80.5}} & \smash{\raisebox{0.4ex}{86.9}} \\
            \bottomrule
        \end{tabular*}
    \end{minipage}
    \vspace{4pt}
    \begin{minipage}[t]{0.49\textwidth}
        \centering
        \includegraphics[width=\linewidth]{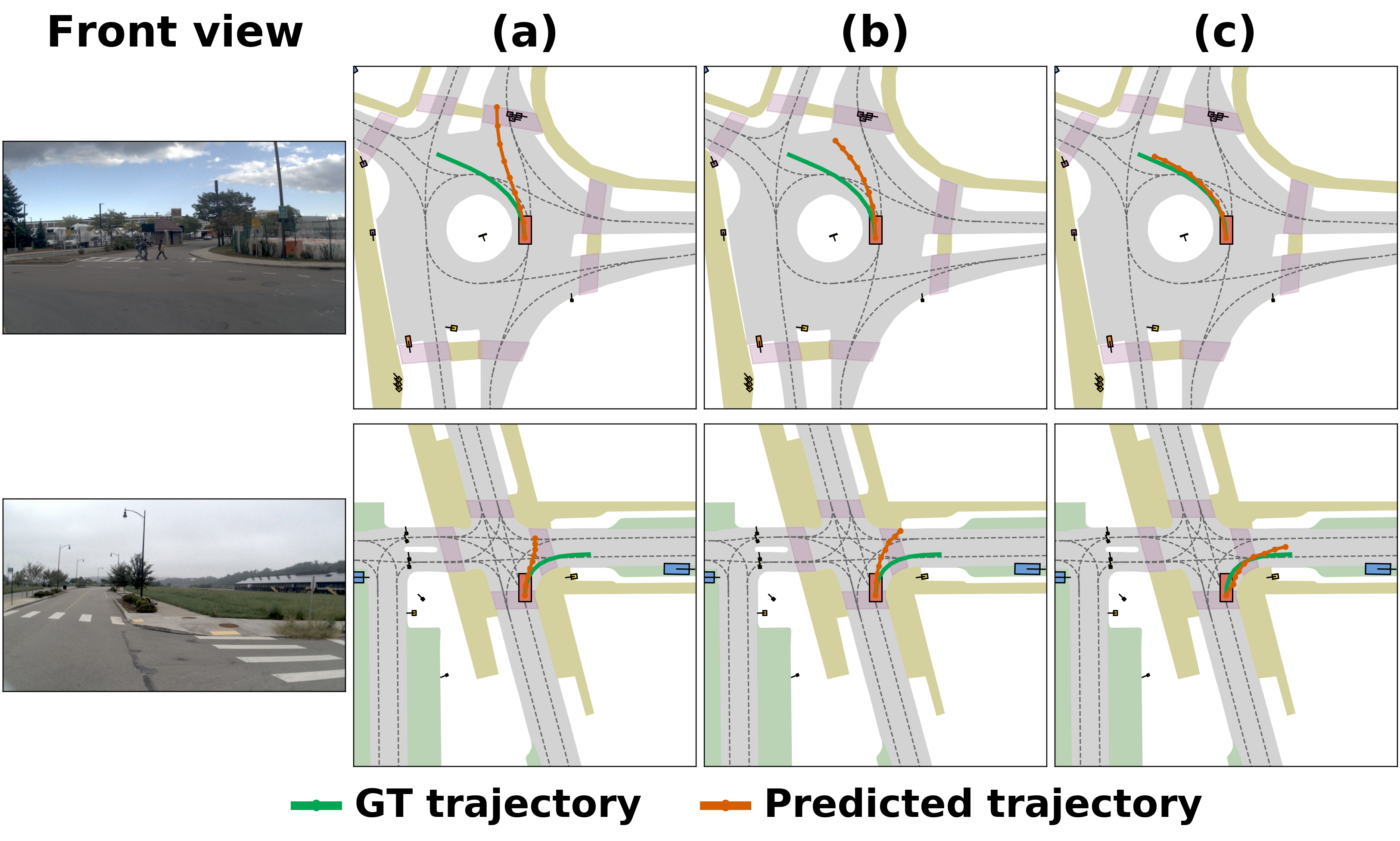}
        \localfigurecaption{\textbf{Qualitative trajectory comparison.}}
        \label{fig:qualitative_navsim}
    \end{minipage}\hfill
    \begin{minipage}[t]{0.49\textwidth}
        \centering
        \includegraphics[width=\linewidth]{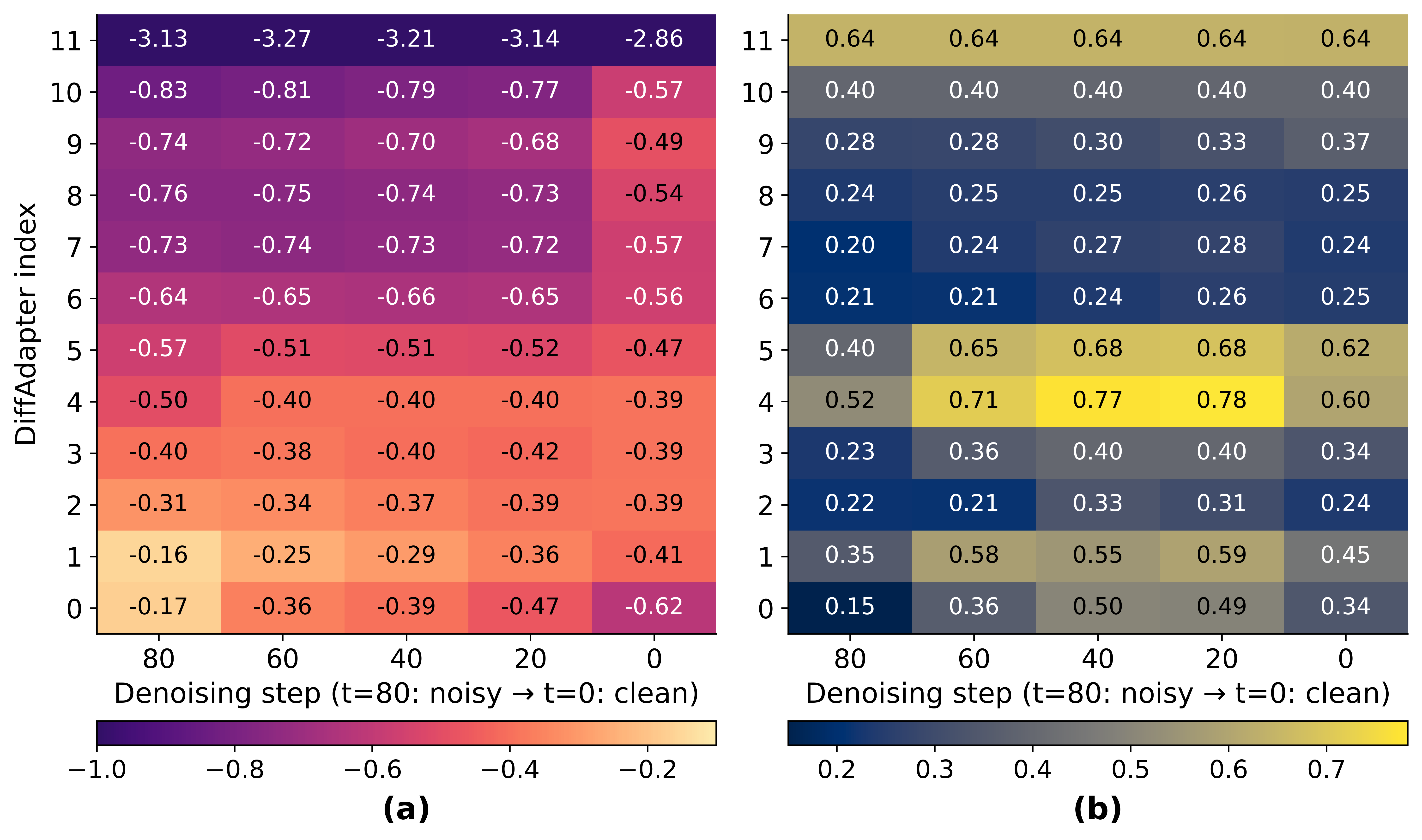}
        \localfigurecaption{\textbf{Layer-wise trajectory refinement.}}
        \label{fig:layerwise_conditioning}
    \end{minipage}
\end{table*}